\documentclass{article}
\usepackage{spconf,amsmath,graphicx}
\usepackage{amsfonts,amssymb}

\title{Scale-aware competition network for palmprint recognition}
\name{Chengrui Gao$^{\dagger}$ \qquad Ziyuan Yang$^{\dagger}$ \qquad Min Zhu$^{\dagger \star}$ \thanks{${\star}$: Corresponding author}\qquad Andrew Beng Jin Teoh$^{\ddagger}$}
  
\address{$^{\dagger}$ College of Computer Science, Sichuan University, Chengdu 610045, China \\
      $^{\ddagger}$ School of Electrical and Electronic Engineering, Yonsei University, Seoul 120749, South Korea}

\begin{document}
%

\maketitle

%
\begin{abstract}

Palmprint biometrics garner heightened attention in palm-scanning payment and social security due to their distinctive attributes. However, prevailing methodologies singularly prioritize texture orientation, neglecting the significant texture scale dimension. We design an innovative network for concurrently extracting intra-scale and inter-scale features to redress this limitation. This paper proposes a scale-aware competitive network (SAC-Net), which includes the Inner-Scale Competition Module (ISCM) and the Across-Scale Competition Module (ASCM) to capture texture characteristics related to orientation and scale. ISCM efficiently integrates learnable Gabor filters and a self-attention mechanism to extract rich orientation data and discern textures with long-range discriminative properties. Subsequently, ASCM leverages a competitive strategy across various scales to effectively encapsulate the competitive texture scale elements. By synergizing ISCM and ASCM, our method adeptly characterizes palmprint features. Rigorous experimentation across three benchmark datasets unequivocally demonstrates our proposed approach's exceptional recognition performance and resilience relative to state-of-the-art alternatives.

\end{abstract}

\section{Introduction}
\label{sec:intro}

Palmprint recognition has gained prominence for identity recognition owing to its elevated accuracy~\cite{Minaee-Bio}. Palmprint imagery encompasses many distinguishing attributes, including principal lines, wrinkles, and an abundance of ridge and minutiae-based features \cite{Jain-A-K-Bio}. Advanced palmprint recognition approaches can be broadly categorized into coding-based, texture-based \cite{sift-based}, subspace-based, and deep learning (DL)-based methodologies.

Among the array of available techniques, coding-based palmprint recognition methods have garnered significant attention due to their commendable precision and free of training. These methods can be categorized into two primary groups: magnitude feature-based and competition-based methodologies. Magnitude feature-based approaches directly encode the magnitude features but are susceptible to lighting variations. In response to this, Kong \textit{et al.}~\cite{compcode} introduced the competitive mechanism in palmprint recognition, introducing the Competitive Code (CompCode). Specifically, CompCode engages in a comparative evaluation of diverse magnitude features across various orientations, encoding the index of the prevailing feature as the ultimate descriptor. Building on the promising performance and robustness of CompCode, researchers have endeavored to devise distinct coding principles for the competition mechanism, resulting in various works such as the Robust Line Orientation Code (RLOC)~\cite{RLOC}, Discriminative Robust Competitive Code (DRCC)~\cite{DRCC}, and Double Orientation Code (DOC) \cite{DOC}. Nevertheless, these methods rely on the feature extractors informed by researchers' a priori knowledge, potentially constraining their capacity for performance enhancement.

Furthermore, researchers have introduced several cutting-edge deep-learning approaches that dispense with manual feature engineering within palmprint recognition. An illustrative instance is the work of Liang \textit{et al}. \cite{Compnet}, wherein they explored utilizing feature ordering information derived from a Convolutional Neural Network (CNN), culminating in what is known as CompNet.

Additionally, Yang \textit{et al}. \cite{co3} endeavored to address this challenge by introducing CO3Net. Their work integrated Coordinate Attention within CO3Net, enabling dynamic emphasis on salient textures using positional information. Nevertheless, CO3Net's assumption that closely related texture information arises primarily from horizontal and vertical coordinates led to a neglect of latent long-range dependencies stemming from diverse orientations. It is worth noting that many existing techniques predominantly emphasize local discriminative orientation, often overlooking the broader context of long-range dependencies and the competitive aspects of texture scale features.

To tackle the challenges above, we introduce a novel network employing Learnable Gabor Filters (LGF) in conjunction with a Multi-Head Self-Attention Mechanism (MSA) and a Scale-Aware Block termed SAC-Net. The primary contributions of our proposed method are as follows: (1) We introduce the innovative Across-Scale Competitive Module (ASCM) designed to extract discriminative scale-related features. (2) We present the Inner-Scale Competition Module (ISCM), which efficiently captures competitive orientation information while accommodating long-range dependencies. (3) By amalgamating ISCM and ASCM, our approach comprehensively characterizes texture features from dual perspectives, encompassing orientation and scale dimensions. (4) Extensive experimentation on three benchmark datasets unequivocally demonstrates that our method outperforms several state-of-the-art alternatives, as evidenced by the substantial results.

\section{Method}
\label{sec:typestyle}

\begin{figure}[!t]
\centerline{\includegraphics[width=\linewidth]{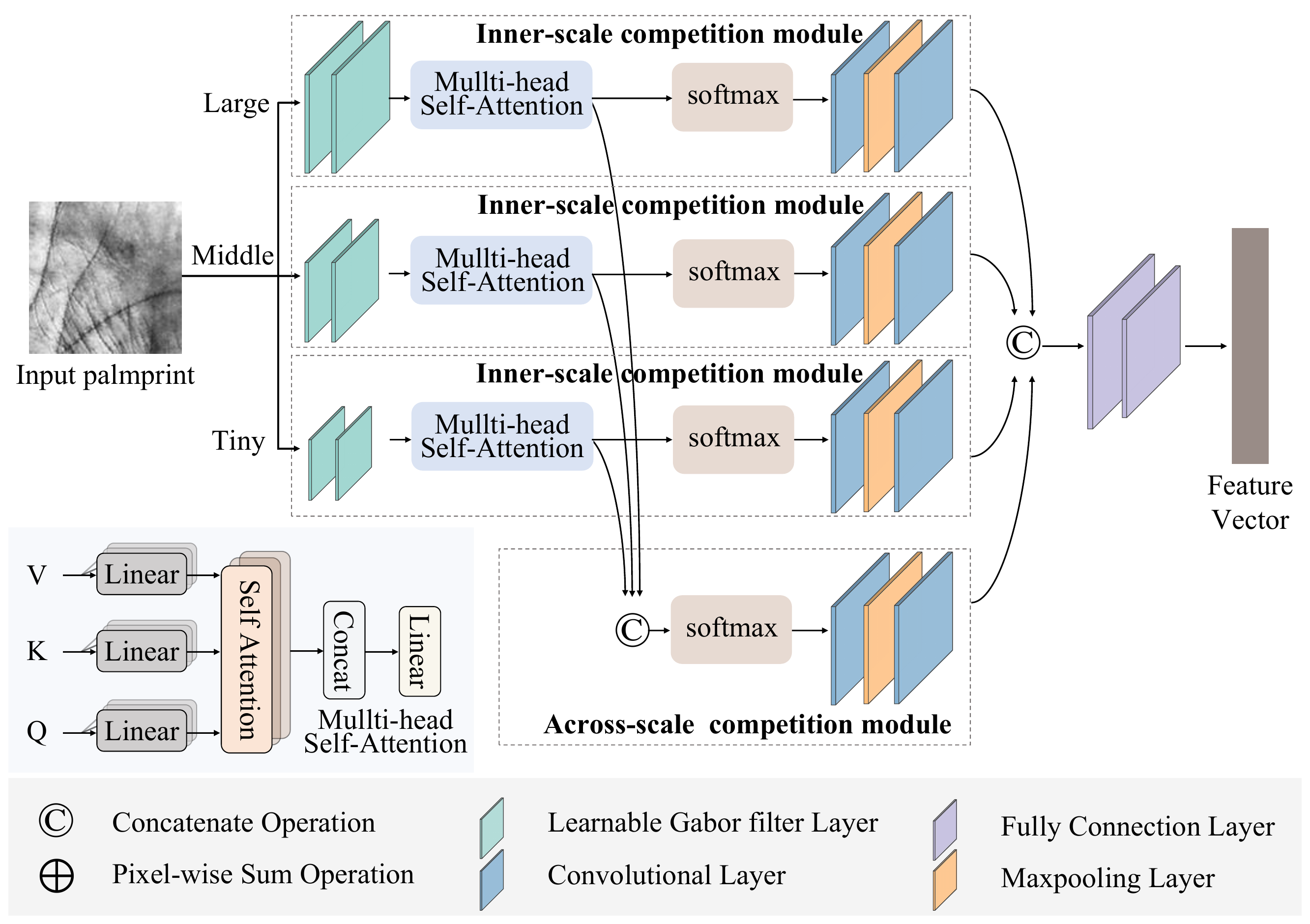}}
\caption{The overview of the proposed SAC-Net.}
\label{framwork}
\vspace{-10pt}
\end{figure}

Fig. \ref{framwork} illustrates the network architecture of SAC-Net, featuring distinct branches tailored for tiny-scale, middle-scale, and large-scale operations (Gabor filter sizes are configured as 7, 17, and 35, correspondingly) to facilitate the extraction of multi-scale texture attributes. Within each branch, the ISCM is deployed to discern textures across various orientations, while the ASCM demonstrates proficient discrimination capabilities across diverse-scale texture attributes. The integration of these components empowers SAC-Net to capture an all-encompassing spectrum of competitive texture features.

\vspace{-10pt}
\subsection{Learnable Gabor Filters}

The Gabor filter, a widely favored descriptor for palmprint recognition, exhibits sensitivity to image edges, affording superior direction and scale selectivity attributes. Furthermore, its resilience to lighting variations renders it highly adaptable to diverse lighting conditions. 
The Gabor function employed in our proposed method adheres to the following formulation:

\vspace{-12pt}
\begin{equation}
g(x, y, \lambda, \theta, \psi, \sigma, \gamma)=e^{ \left(-\frac{x^{\prime 2}+\gamma^2 y^{\prime 2}}{2 \sigma^2}\right)} \cos \left(\frac{2 \pi x^{\prime}}{\lambda}+\psi\right),
\end{equation}
where $\begin{gathered}x^{\prime}=x \cos \theta+y \sin \theta\end{gathered}$ and $\begin{gathered}y^{\prime}=-x \sin \theta+y \cos \theta \end{gathered}$. $(x, y)$ represents the pixel coordinate index.
$\lambda$ and $\psi$ are the wavelength and phase parameters of the cosine function in the Gabor filter, respectively.
$\theta$ is the orientation of the Gabor filter.
$\gamma$ is the spatial aspect ratio of the Gaussian function, which determines the shape of the Gabor function.
$\sigma$ is the standard deviation.

To circumvent the unanticipated heterogeneity stemming from manually crafted Gabor filters, we opt for Learnable Gabor Filters (LGF) as texture extraction tools, which facilitate learning optimal parameters to extract texture features.

\begin{figure}[!t]
	\centering
	\begin{minipage}[t]{0.3\columnwidth}
	\centering
	\includegraphics[width=\columnwidth]{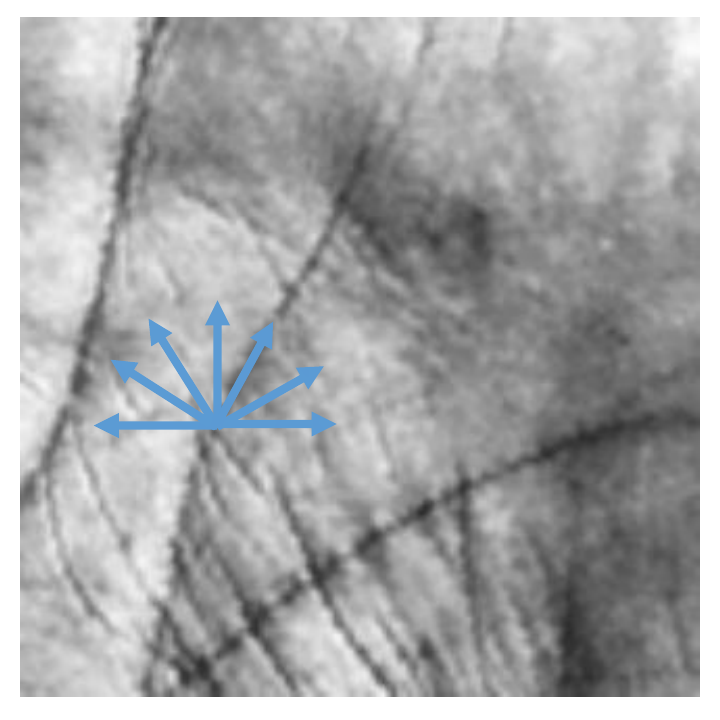}
	\centerline{(a) ISCM}
	\end{minipage}
	\begin{minipage}[t]{0.3\columnwidth}
	\includegraphics[width=\columnwidth]{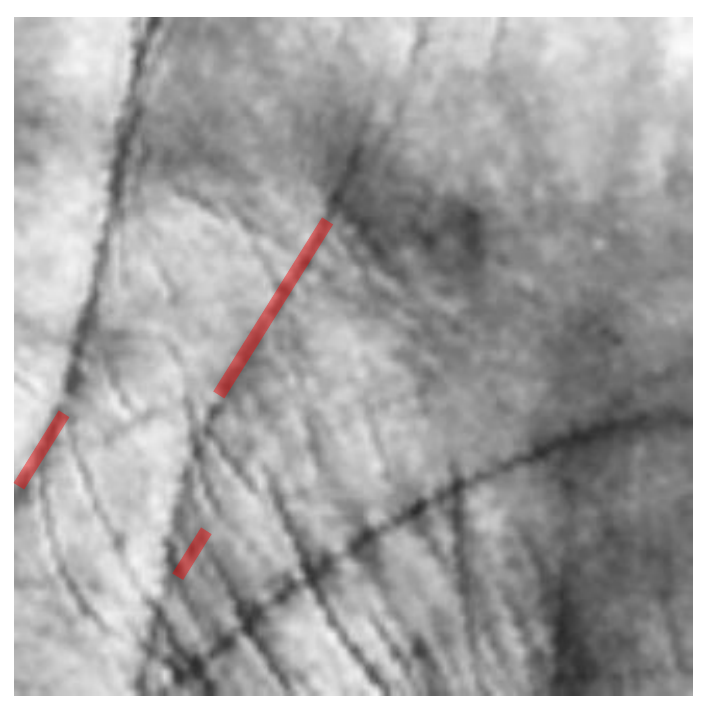}
	\centerline{(b) ASCM}
	\end{minipage}

\caption{Illustrations of our method for extracting texture feature}
\vspace{-10pt}
\label{fig2}
\end{figure}

\subsection{Competition Module}

Traditional competitive mechanisms prioritize evaluating responses across various channel dimensions, selecting the maximum response as the winner and utilizing the winning orientation index as a feature. This approach primarily concentrates on local discriminative orientational characteristics to pinpoint the optimal texture orientation. However, it neglects the texture scale, discarding all other attributes except for the winner and overlooking potential long-range dependency connections.

To transcend the limitations inherent in competitive mechanisms, we introduce a novel competition module that extends competition to encompass both orientational and texture scale dimensions. This expanded scope allows for a more comprehensive assimilation of orientational and texture-related features. As depicted in Fig. \ref{framwork}, this module comprises the ISCM and the ASCM. In pursuit of capturing potentially valuable insights, we employ the Softmax-based Competitive Coding (SCC) for competition feature extraction \cite{Compnet}, preserving all responses from the LGF. SCC utilizes the Softmax function to derive ordinal relationships as features, encompassing a broader spectrum of information beyond solely the winner's index.

\vspace{-9pt}
\subsubsection{Inner-Scale Competition Module}

We introduce the ISCM within each branch of SAC-Net. Initially, the module captures local texture details via LGF. Subsequently, an MSA module is applied to model extensive-range dependencies, capturing overarching texture characteristics. Additionally, the network employs SCC to emphasize discriminative texture attributes along the texture orientation. This procedure facilitates the elucidation of the sequential relationships among texture features derived from different descriptors. Taking the large-scale branch as an example, the procedure can be expressed as follows:

\vspace{-10pt}
\begin{equation}
\begin{aligned}
\mathbf{F}^{ls}_{LGF}=G_2(G_1\left(\mathbf{X}\right)),
\end{aligned}
\end{equation}
where $\mathbf{X} \in \mathbb{R}^{b \times c\times h\times w} $ be the input image, $b, c, h, w$ refer to the batch size, channel, height and weight of the input, respectively. $G_1(\cdot)$ and $G_2(\cdot)$ are the processes of each branch's first and second LGF layers. $\mathbf{F}^{ls}_{LGF}$ is the local competitive orientation feature of the palmprint image.

\begin{equation}
\begin{aligned}
\mathbf{F}^{ls}_{MSA}=Norm(MultiHead\left(\mathbf{F}^{ls}_{LGF}\right)),
\end{aligned}
\end{equation}
where $MultiHead(\cdot)$ is the multi-head self-attention operation. $Norm(\cdot)$  the normalization operation. $\mathbf{F}^{ls}_{MSA}$ represents the direction ordering features with the long-range dependency relationship.
\vspace{-7pt}
\begin{equation}
\begin{aligned}
\mathbf{F}^{ls}_{inner}=softmax\left(\mathbf{F}^{ls}_{MSA} \right),
\end{aligned}
\end{equation}
where $softmax(\cdot)$ denotes the competition features extraction process along orientations. $\mathbf{F}^{ls}_{inner}$ is the competition information along the inner-scale feature orientation.

\vspace{-5pt}
\subsubsection{Across-Scale Competition Module}

We introduce the ASCM, accompanied by a strategy for comparing feature scales across different branches. The fundamental premise of this cross-scale competition lies in the assumption that when the texture size aligns with the filter size, the response attains its maximum value, which is deemed scale feature information. Our initial step involves capturing multi-scale global features spanning the large-scale, middle-scale, and tiny-scale branches, which can be expressed as:

\begin{equation}
\mathbf{F}_{MSA}= Concat(\mathbf{F}^{ls}_{MSA}, \mathbf{F}^{ms}_{MSA}, \mathbf{F}^{ts}_{MSA}),
\end{equation}
where the $\mathbf{F}^{ls}_{MSA}$, $\mathbf{F}^{ms}_{MSA}$, $\mathbf{F}^{ts}_{MSA}$ represent the direction ordering features with the long-range dependency relationship from large-scale, middle-scale and tiny-scale branches, respectively. $Concat$ is the concatenate operation.

Then, we can extract the ordering of discriminant scale features by applying the softmax function:
\vspace{-5pt}

\begin{equation}
\mathbf{F}_{across}=softmax\left(\mathbf{F}_{MSA} \right),
\end{equation}
where $\mathbf{F}_{across}$ is the competition information along the texture scale size.

The palmprint ROIs are depicted in Fig. \ref{fig2}. In Fig. \ref{fig2}(a), we visualize competitive features acquired through the ISCM along the texture orientation. Conversely, in Fig. \ref{fig2}(b), competitive features captured by the ASCM for distinct texture scales are illustrated. Fig. \ref{fig2}(b) reveals the diverse texture sizes in palmprint images. The ability to capture competitive relationships among these varied textures enhances the representation of texture information, a pivotal aspect in palmprint recognition.

\section{Experiments}
\label{sec:majhead}

We perform a series of experiments to evaluate our proposed method on three publicly available palmprint datasets: IITD~\cite{kumar2010personal}, PolyU~\cite{zhang2003online}, and multi-spectral~\cite{zhang2009online}. The contact devices acquire PolyU and multi-spectral datasets, and IITD are contactless palmprint datasets. The multi-spectral dataset contains four spectral bands, including red, green, blue, and NIR. In this paper, we construct the loss by combining the cross-entropy and contrastive loss~\cite{co3}. The training involves utilizing the Adam optimizer with a fixed learning rate of 0.0003 for all networks. Our network is implemented using the PyTorch framework and trained on a single NVIDIA RTX 3090 GPU.

\begin{figure}[t]
	\centering
    \begin{minipage}[t]{0.48\columnwidth}
	\centering
	\includegraphics[width=\columnwidth]{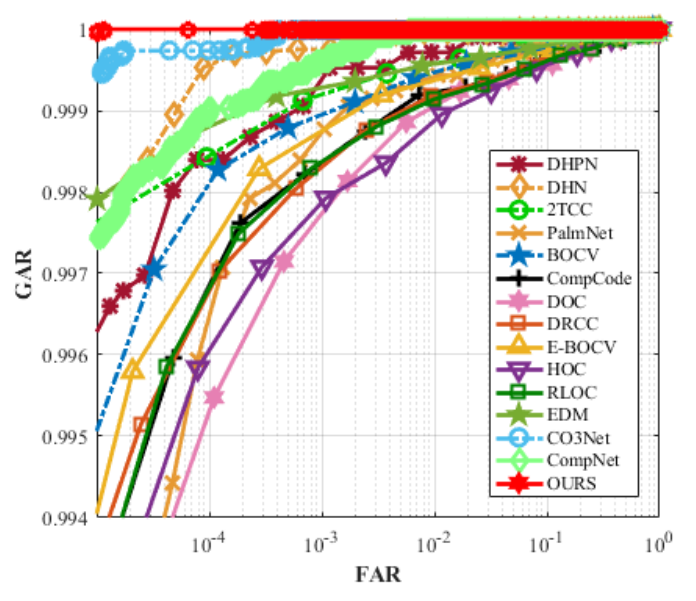}
	\centerline{(a) PolyU}
	\end{minipage}
	\begin{minipage}[t]{0.48\columnwidth}
	\includegraphics[width=\columnwidth]{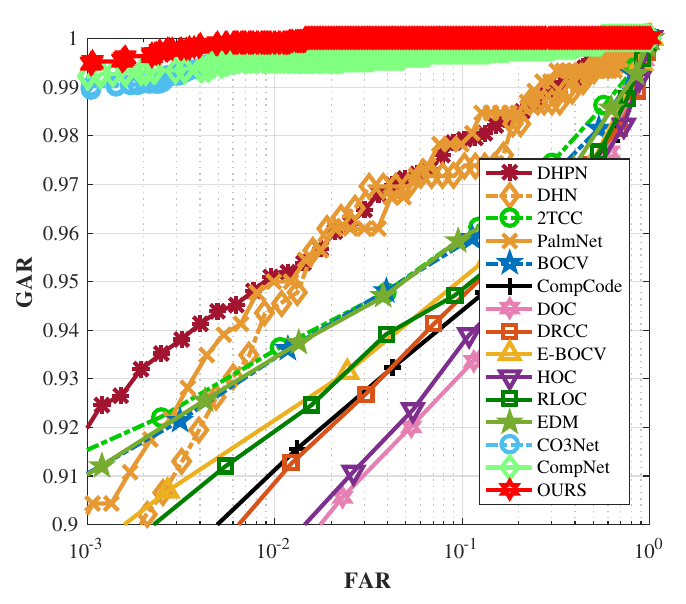}
	\centerline{(b) IITD}
	\end{minipage}
	\begin{minipage}[t]{0.48\columnwidth}
	\centering
	\includegraphics[width=\columnwidth]{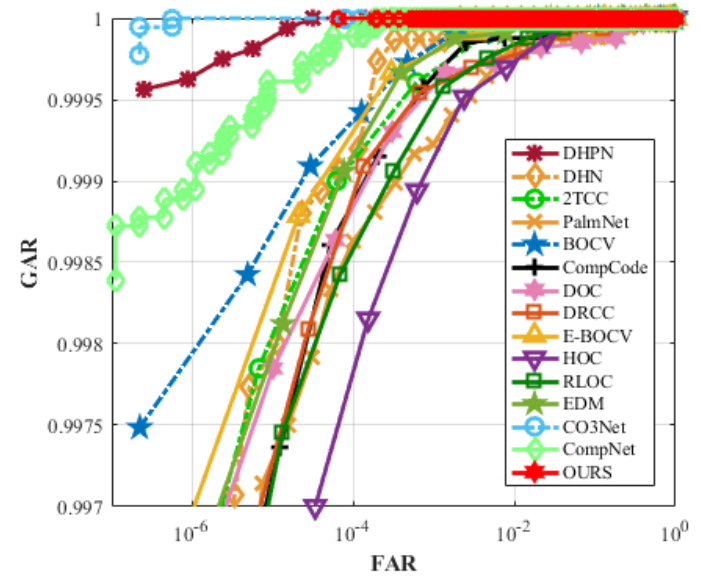}
	\centerline{(c) NIR}
	\end{minipage}
	\begin{minipage}[t]{0.48\columnwidth}
	\includegraphics[width=\columnwidth]{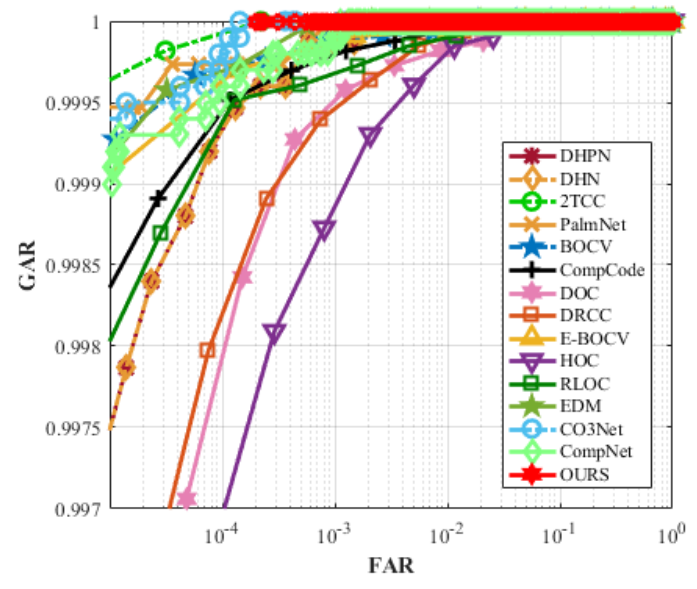}
	\centerline{(d) Red}
	\end{minipage}
    \begin{minipage}[t]{0.48\columnwidth}
	\includegraphics[width=\columnwidth]{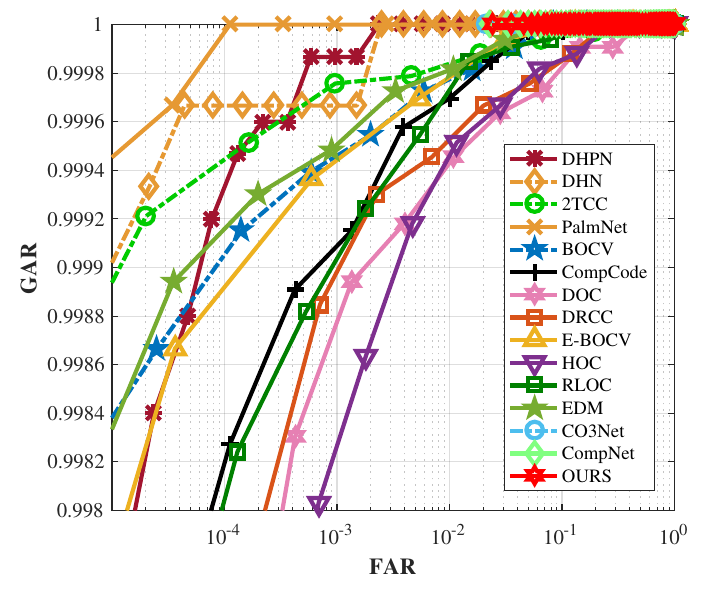}
	\centerline{(e) Green}
	\end{minipage}
    \begin{minipage}[t]{0.48\columnwidth}
	\includegraphics[width=\columnwidth]{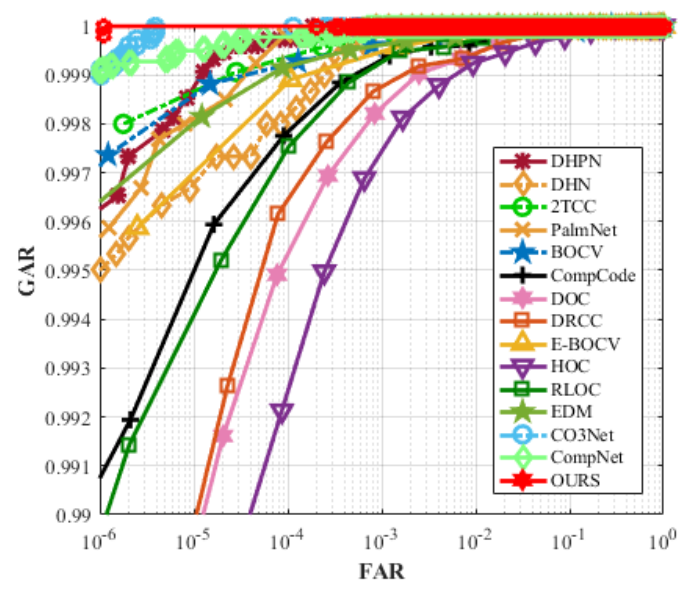}
	\centerline{(f) Blue}
	\end{minipage}
\caption{The ROC curves of the proposed and its competing methods across all datasets.}
\label{roc}
\label{fig2}
\end{figure}

\begin{table}[t]
\centering
\caption{EERs on IITD and PolyU datasets (\%)}
\label{eer1}
\begin{tabular}{ccc}
\hline
            & IITD            & PolyU           \\ \hline
CompCode \cite{compcode}    & 5.50          & 0.1200          \\
RLOC \cite{RLOC}         & 5.00          & 0.1300          \\
HOC \cite{HOC}         & 6.55          & 0.1600          \\
DOC \cite{DOC}          & 6.20          & 0.1800         \\
DRCC \cite{DRCC}      & 5.42          & 0.1800          \\
BOCV \cite{guo2009palmprint}        & 4.56          & 0.0813          \\
E-BOCV \cite{zhang2012fragile}       & 4.65          & 0.0995          \\
EDM \cite{yang2020extreme}        & 4.56          & 0.0609          \\
DHN \cite{wu2021palmprint}        & 4.30          & 0.0372          \\
DHPN \cite{zhong2018palm}           & 3.73          & 0.0320          \\
PalmNet \cite{genovese2019palmnet}       & 4.20          & 0.1110          \\
2TCC \cite{yang2023multi}         & 5.94          & 0.0834          \\
CompNet \cite{Compnet}      & 0.54          & 0.0556          \\
CO3Net \cite{co3}      & 0.47          & 0.0220          \\
\textbf{OURS} & \textbf{0.25} & \textbf{0.0012} \\ \hline
\end{tabular}
\vspace{-10pt}
\end{table}

\vspace{-15pt}
\begin{table}[]
\centering
\caption{EERs on multi-spectral datasets (\%)}
\label{eer2}
\begin{tabular}{ccccc}
\hline
            & Red             & Green           & Blue            & NIR             \\ \hline
CompCode \cite{compcode}    & 0.0357          & 0.1100          & 0.0911          & 0.0579          \\
RLOC \cite{RLOC}         & 0.0444          & 0.0855          & 0.0799          & 0.0629          \\
HOC \cite{HOC}          & 0.1000          & 0.1600          & 0.1800          & 0.0839          \\
DOC \cite{DOC}         & 0.0584          & 0.1200          & 0.1300          & 0.0501          \\
DRCC \cite{DRCC}        & 0.0660          & 0.0927          & 0.1100          & 0.0563          \\
BOCV \cite{guo2009palmprint}        & 0.0164          & 0.0442          & 0.0358          & 0.0261          \\
E-BOCV \cite{zhang2012fragile}      & 0.0216          & 0.0616          & 0.0599          & 0.0315          \\
EDM \cite{yang2020extreme}         & 0.0206          & 0.0703          & 0.0473          & 0.0363          \\
DHN \cite{wu2021palmprint}           & 0.0380          & 0.0304          & 0.0403          & 0.0233          \\
DHPN \cite{zhong2018palm}         & 0.0369          & 0.0352          & 0.0213          & 0.0020          \\
PalmNet \cite{genovese2019palmnet}      & 0.0366          & 0.0087          & 0.0178          & 0.0871          \\
2TCC \cite{yang2023multi}         & 0.0185          & 0.0375          & 0.0405          & 0.0322          \\
CompNet \cite{Compnet}      & \textbf{0.0000}          & 0.0055          & 0.0173          & 0.0025          \\
CO3Net \cite{co3}       & \textbf{0.0000}          & 0.0009          & 0.0004          & 0.0005          \\
\textbf{OURS} & \textbf{0.0000} & \textbf{0.0000} & \textbf{0.0001} & \textbf{0.0000} \\ \hline
\end{tabular}
\end{table}

\subsection{Verification Experiments}
\label{sssec:subsubhead}

We evaluate the verification performance using Receiver Operating Characteristic (ROC) curves and the Equal Error Rate (EER), both derived from Genuine Acceptance Rate (GAR) and False Accept Rate (FAR). In ROC, superior performance is indicated by proximity to the top. Fig. \ref{roc} presents ROC curves for SAC-Net and its competing counterparts, consistently positioning SAC-Net at the apex, outperforming the prior fourteen methods across all datasets. This underscores SAC-Net's superior performance across diverse palmprint acquisition devices.

Tables \ref{eer1}-\ref{eer2} present EERs for SAC-Net versus other methods where a lower EER indicates better performance. The results reveal markedly lower EERs for SAC-Net across all datasets. Compcode, CO3Net, and SAC-Net consistently outperform their counterparts, signifying the robust and discriminative nature of competitive features in palmprint recognition. SAC-Net's notable performance can be attributed to its scale-aware mechanism, facilitating the comprehensive capture of competitive texture attributes. SAC-Net's remarkable performance on multispectral datasets is particularly noteworthy, achieving 0\% EERs on Red, Green, and NIR datasets.

\vspace{-10pt}
\subsection{Ablation Study}
\label{sssec:subsubhead}

We first study different branch scales' impact on the verification performance. Table \ref{branch} reveals that, on the IITD dataset, the middle-scale branch exhibits superior performance compared to other single-scale branches. Additionally, performance gains are observed with increased multiscale texture extractors, signifying the complementary nature of features across different scales. Balancing performance and computational complexity, we opt for a configuration of three-scale texture extractors in SAC-Net.

\begin{table}[t]
\centering
\caption{Ablation study for branch's scale on IITD dataset.}
\label{branch}
\begin{tabular}{cccc}
\hline
Tiny scale & Middle scale & Large scale & EER(\%)             \\
\hline
$\checkmark$          & $\times$            & $\times$           & 3.0700          \\
$\times$          & $\checkmark$            & $\times$           & 0.6455          \\
$\times$          & $\times$            & $\checkmark$           & 1.2319          \\
$\checkmark$          & $\checkmark$            & $\times$           & 0.5435          \\
$\times$          & $\checkmark$            & $\checkmark$          & 0.4710          \\
$\checkmark$          & $\checkmark$            & $\checkmark$           & \textbf{0.2536}\\ \hline
\end{tabular}
\vspace{-10pt}
\end{table}

We conducted an additional ablation experiment on the IITD dataset to assess the efficacy of the two main modules within our SAC-Net. We established a baseline using SAC-Net devoid of the ISCM and ASCM. Subsequently, we introduce the ISCM, ASCM, and both modules to evaluate their respective contributions. The outcomes, as presented in Table \ref{module}, reveal that the amalgamation of ISCM and ASCM modules leads to performance improvement. Furthermore, our scale-aware mechanism substantially enhances the performance of the baseline configuration.

\begin{table}[t]
\centering
\caption{Ablation study for ISCM and ASCM on IITD dataset.}
\label{module}
\begin{tabular}{ccc}
\hline
ASCM & ISCM & IITD/EER(\%) \\ \hline
$\times$    & $\times$    & 0.6522   \\
$\times$    & $\checkmark$    & 0.4348   \\
$\checkmark$    & $\times$    & 0.3262   \\
$\checkmark$    & $\checkmark$    & \textbf{0.2536}   \\ \hline
\end{tabular}
\vspace{-10pt}
\end{table}

\section{Conlusion}
\label{sec:majhead}

We introduce SAC-Net, a scale-aware network designed for palmprint biometrics by harnessing comprehensive competitive orientation and scale features by integrating ISCM and ASCM. ISCM adeptly captures discriminative orientation characteristics, incorporating global palmprint features. Subsequently, ASCM is employed to emphasize competitive scale attributes. SAC-Net exhibits applicability in contact-based and contactless palmprint verification scenarios, extending its utility to palmprint images acquired under various spectral bands. Our experimental results consistently demonstrate SAC-Net's superiority over traditional encoding and deep-learning-based approaches across three distinct datasets.

\newpage

\bibliographystyle{IEEEbib}
\bibliography{refs}

\begin{thebibliography}{10}

\bibitem{Minaee-Bio}
Shervin Minaee, Amirali Abdolrashidi, Hang Su, Mohammed Bennamoun, and David Zhang,
\newblock ``Biometrics recognition using deep learning: A survey,''
\newblock {\em Artificial Intelligence Review}, pp. 1--49, 2023.

\bibitem{Jain-A-K-Bio}
Anil~K Jain and Ajay Kumar,
\newblock ``Biometric recognition: an overview,''
\newblock {\em Second Generation Biometrics: The Ethical, Legal and Social Context}, pp. 49--79, 2012.

\bibitem{compcode}
AW-K Kong and David Zhang,
\newblock ``Competitive coding scheme for palmprint verification,''
\newblock in {\em Proceedings of the 17th International Conference on Pattern Recognition.} IEEE, 2004, vol.~1, pp. 520--523.

\bibitem{RLOC}
Wei Jia, De-Shuang Huang, and David Zhang,
\newblock ``Palmprint verification based on robust line orientation code,''
\newblock {\em Pattern Recognition}, vol. 41, no. 5, pp. 1504--1513, 2008.

\bibitem{DRCC}
Yong Xu, Lunke Fei, Jie Wen, and David Zhang,
\newblock ``Discriminative and robust competitive code for palmprint recognition,''
\newblock {\em IEEE Transactions on Systems, Man, and Cybernetics: systems}, vol. 48, no. 2, pp. 232--241, 2016.

\bibitem{DOC}
Lunke Fei, Yong Xu, Wenliang Tang, and David Zhang,
\newblock ``Double-orientation code and nonlinear matching scheme for palmprint recognition,''
\newblock {\em Pattern Recognition}, vol. 49, pp. 89--101, 2016.

\bibitem{HOC}
Lunke Fei, Yong Xu, and David Zhang,
\newblock ``Half-orientation extraction of palmprint features,''
\newblock {\em Pattern Recognition Letters}, vol. 69, pp. 35--41, 2016.

\bibitem{Compnet}
Xu~Liang, Jinyang Yang, Guangming Lu, and David Zhang,
\newblock ``Compnet: Competitive neural network for palmprint recognition using learnable gabor kernels,''
\newblock {\em IEEE Signal Processing Letters}, vol. 28, pp. 1739--1743, 2021.

\bibitem{co3}
Ziyuan Yang, Wenjun Xia, Yifan Qiao, Zexin Lu, Bob Zhang, Lu~Leng, and Yi~Zhang,
\newblock ``Co3net: Coordinate-aware contrastive competitive neural network for palmprint recognition,''
\newblock {\em IEEE Transactions on Instrumentation and Measurement}, 2023.

\bibitem{sift-based}
Xiangqian Wu, Qiushi Zhao, and Wei Bu,
\newblock ``A sift-based contactless palmprint verification approach using iterative ransac and local palmprint descriptors,''
\newblock {\em Pattern Recognition}, vol. 47, no. 10, pp. 3314--3326, 2014.

\bibitem{zhang2003online}
David Zhang, Wai-Kin Kong, Jane You, and Michael Wong,
\newblock ``Online palmprint identification,''
\newblock {\em IEEE Transactions on Pattern Analysis and Machine Intelligence}, vol. 25, no. 9, pp. 1041--1050, 2003.

\bibitem{kumar2010personal}
Ajay Kumar and Sumit Shekhar,
\newblock ``Personal identification using multibiometrics rank-level fusion,''
\newblock {\em IEEE Transactions on Systems, Man, and Cybernetics, Part C (Applications and Reviews)}, vol. 41, no. 5, pp. 743--752, 2010.

\bibitem{zhang2009online}
David Zhang, Zhenhua Guo, Guangming Lu, Lei Zhang, and Wangmeng Zuo,
\newblock ``An online system of multispectral palmprint verification,''
\newblock {\em IEEE Transactions on Instrumentation and Measurement}, vol. 59, no. 2, pp. 480--490, 2009.

\bibitem{guo2009palmprint}
Zhenhua Guo, David Zhang, Lei Zhang, and Wangmeng Zuo,
\newblock ``Palmprint verification using binary orientation co-occurrence vector,''
\newblock {\em Pattern Recognition Letters}, vol. 30, no. 13, pp. 1219--1227, 2009.

\bibitem{zhang2012fragile}
Lin Zhang, Hongyu Li, and Junyu Niu,
\newblock ``Fragile bits in palmprint recognition,''
\newblock {\em IEEE Signal processing letters}, vol. 19, no. 10, pp. 663--666, 2012.

\bibitem{yang2020extreme}
Ziyuan Yang, Lu~Leng, and Weidong Min,
\newblock ``Extreme downsampling and joint feature for coding-based palmprint recognition,''
\newblock {\em IEEE Transactions on Instrumentation and Measurement}, vol. 70, pp. 1--12, 2020.

\bibitem{wu2021palmprint}
Tengfei Wu, Lu~Leng, Muhammad~Khurram Khan, and Farrukh~Aslam Khan,
\newblock ``Palmprint-palmvein fusion recognition based on deep hashing network,''
\newblock {\em IEEE Access}, vol. 9, pp. 135816--135827, 2021.

\bibitem{zhong2018palm}
Dexing Zhong, Shuming Liu, Wenting Wang, and Xuefeng Du,
\newblock ``Palm vein recognition with deep hashing network,''
\newblock in {\em Proceedings of the 1st Pattern Recognition and Computer Vision}. Springer, 2018, pp. 38--49.

\bibitem{genovese2019palmnet}
Angelo Genovese, Vincenzo Piuri, Konstantinos~N Plataniotis, and Fabio Scotti,
\newblock ``Palmnet: Gabor-pca convolutional networks for touchless palmprint recognition,''
\newblock {\em IEEE Transactions on Information Forensics and Security}, vol. 14, no. 12, pp. 3160--3174, 2019.

\bibitem{yang2023multi}
Ziyuan Yang, Lu~Leng, Tengfei Wu, Ming Li, and Jun Chu,
\newblock ``Multi-order texture features for palmprint recognition,''
\newblock {\em Artificial Intelligence Review}, vol. 56, no. 2, pp. 995--1011, 2023.

\end{thebibliography}

\end{document}